\documentclass[10pt,twocolumn,letterpaper]{article}

\usepackage[pagenumbers]{cvpr} 
\usepackage{booktabs}
\usepackage{multirow}
\usepackage{graphicx}
\usepackage{xcolor}
\usepackage{colortbl}
\usepackage{bbding}
\definecolor{oursgray}{gray}{0.92} 

\usepackage{cuted} 

\usepackage{currfile} 

\usepackage{caption} 

\newcommand{\nothing}[1]{{}}

\newcommand{\TODO}[1]{\textbf{\color{red}[TODO: #1]}}

\renewcommand{\paragraph}[1]{\vspace{.5em}\noindent\textbf{#1}}

\definecolor{cvprblue}{rgb}{0.21,0.49,0.74}
\usepackage[pagebackref,breaklinks,colorlinks,allcolors=cvprblue]{hyperref}

\usepackage[normalem]{ulem}

\def\confName{CVPR}
\def\confYear{2027}

\newcommand{\acronym}{CIDER\xspace}

\title{Causal Diffusion for Extreme Super-Resolution in Interactive Games}

\author{
Christina Zhang\textsuperscript{1,*} \hspace{1em}
Jenna Kang\textsuperscript{1,*} \hspace{1em}
Colin Groth\textsuperscript{1} \\
Akshay Jindal\textsuperscript{2} \hspace{1em}
Stavros Diolatzis\textsuperscript{2} \hspace{1em}
Qi Sun\textsuperscript{1} \\
\vspace{0.5em}
\textsuperscript{1}New York University \hspace{1em}
\textsuperscript{2}Intel Corporation \\
{\tt\small \{cz3641, jennakang, cg4932, qisun\}@nyu.edu} \\
{\tt\small \{akshay.jindal, stavros.diolatzis\}@intel.com} \\
\vspace{0.3em}
{\small \textsuperscript{*}Equal contribution}
}

\begin{document}
\maketitle
%
\begin{strip}
\centering
\includegraphics[width=\linewidth]{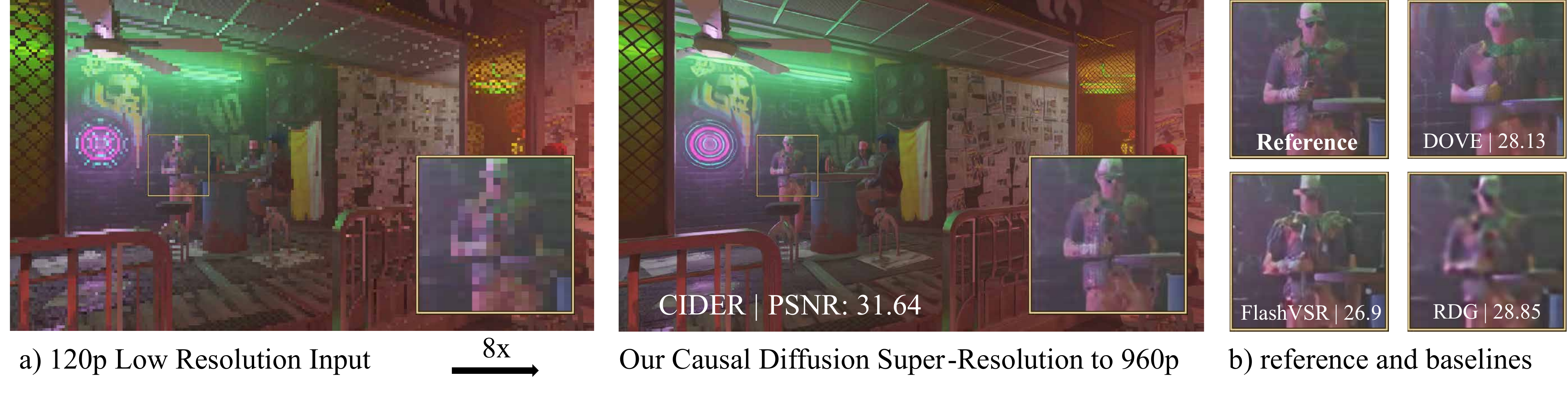}
\captionof{figure}{a) We introduce \acronym, a super-resolution method which utilizes causal diffusion to enable modern video game upscaling from low-resolution inputs. Our method can infer the complex geometries and textures of the scene from the ultra-low-resolution input while undoing the strong aliasing artifacts present. b) In this ultra-low input resolution regime the baselines fail to reconstruct the scene, suffering from spatial and temporal artifacts.}
\captionsetup{hypcap=true}
\label{fig:teaser}%
\end{strip}

\begin{abstract}
Existing game super-resolution methods typically operate at moderate upscaling factors, struggling to recover fine visual details from extremely low-resolution inputs (e.g., 120p). At such resolutions, super-resolution shifts from reconstructing existing information toward synthesizing missing visual details. While video diffusion models offer powerful generative priors for this challenging task, their dependence on future frames limits their applicability to interactive games. We present \acronym, a causal diffusion framework for extreme game super-resolution that synthesizes high-frequency details from current and past low-resolution frames, guided by G-buffers (depth and motion vectors) to preserve scene structure and temporal consistency. We collect data from two AAA games (Cyberpunk 2077 and Black Myth: Wukong). To reduce inference load, we develop a lightweight latent projection encoder tailored to ultra-low-resolution inputs, fine-tune a decoder (TAE) on our high-resolution outputs, and distill the diffusion model to reach interactive frame rates. Our experiments demonstrate high-fidelity 8$\times$ upscaling from 120p to 960p at around 13 FPS, outperforming state-of-the-art game upscalers in visual quality and showing potential for 16$\times$ magnification.

\end{abstract}
    
\section{Introduction}
\label{sec:intro}
High-quality AAA games often require intensive rendering pipelines with ray tracing, which significantly limits their performance on budget GPUs or edge devices. Neural super-resolution has become an integral component of modern game rendering, enabling high-resolution visual experiences from low-resolution run-time shading. Commercial solutions such as DLSS \cite{nvidia_dlss}, FSR \cite{amd_fsr}, and XeSS \cite{intel_xess} have demonstrated impressive reconstruction quality at moderate upscaling factors. However, \emph{how far can we push game super-resolution when the input resolution becomes extremely low?} Existing methods typically operate at scaling factors of up to 4$\times$ per spatial dimension \cite{zhuang2026flashvsr}, leaving extreme upscaling ratios such as 8$\times$ or even 16$\times$ largely unexplored. At these scales, recovering high-fidelity details becomes fundamentally more challenging: an input frame at 120p contains only 1/64 of the pixels of its 960p output, with much of the texture, fine geometry, and high-frequency visual information missing (Fig.~\ref{fig:teaser}a, left).

Such extreme conditions shift super-resolution from primarily reconstructing existing information toward synthesizing missing visual details. Conventional game upscalers, often built on convolutional or other regression-based architectures \cite{xiao2020neural, guo2021extranet, thomas2022temporally}, struggle when the input provides insufficient visual evidence. Diffusion models, in contrast, have demonstrated a remarkable ability to generate realistic, high-frequency content through learned generative priors  \cite{rombach2022ldm, yue2023resshift, yu2024supir}. This capability makes diffusion particularly promising for extreme game super-resolution, where recovering details directly from the low-resolution input is severely underconstrained. However, generative models also introduce the risk of hallucinating scene content, making fidelity to the original game rendering a critical requirement. As illustrated in \cref{fig:results}, extreme super-resolution demands not only plausible detail synthesis but also faithful preservation of scene geometry and appearance.

Applying video diffusion models to interactive games presents an additional challenge: future frames are unavailable. Unlike offline video super-resolution, where models can exploit bidirectional temporal information, interactive games must generate each frame as new gameplay unfolds. A practical approach must therefore operate causally while maintaining temporal consistency and preserving the scene's geometry and appearance under dynamic motion. These requirements rule out the use of conventional offline video enhancement for extreme game super-resolution. Additionally, current real-time super-resolution approaches using diffusion models mostly generate chunks of frames at a time, which is not applicable to gaming as it leads to high latency. Therefore, super-resolution for gaming must focus on low latency for single frame generation instead of an amortized FPS metric. 

To break these limitations, we present \underline{\textbf{\acronym}}, \underline{\textbf{C}}ausal \underline{\textbf{I}}nteractive \underline{\textbf{D}}iffusion for \underline{\textbf{E}}xtreme \underline{\textbf{R}}esolution. By adapting the pretrained video diffusion model \cite{chen2025dove}, \acronym synthesizes high-frequency details using only current and previous low-resolution frames, without relying on future observations nor accumulating generative errors. Similar to prior work \cite{zheng2025rdg}, we further incorporate in-engine G-buffers, including depth and motion vectors, to guide the generative process with explicit geometric and motion information. Together with game-specific fine-tuning, these conditioning signals help preserve scene structure and temporal consistency while reducing generative artifacts (see \cref{fig:teaser}a, right).

To make diffusion-based super-resolution suitable for interactive applications, we further optimize the inference pipeline. Specifically,
we replace the computationally expensive VAE encoder in the diffusion transformer with a lightweight latent projection designed for ultra-low-resolution inputs and fine-tune a decoder (TAE) for high-resolution reconstruction. We further accelerate inference through student model distillation, temporally local causal attention with key-value caching, and optimized fused attention kernels. Together, these modifications allow our framework to process extreme super-resolution tasks at approximately 0.075 seconds per frame on an NVIDIA RTX Pro 6000 Blackwell GPU, demonstrating the feasibility of causal diffusion-based generation under interactive constraints.

We evaluate \acronym on two graphically demanding AAA games, Cyberpunk 2077 and Black Myth: Wukong. Our method achieves high-quality 8$\times$ super-resolution from 120p to 960p, substantially outperforming existing game upscalers in visual quality under such extreme input conditions. We further investigate the potential for 16$\times$ magnification and reconstruction from inputs as low as 60p.
In summary, we make the following main contributions:
\begin{itemize}
    \item We demonstrate the feasibility of diffusion-based super-resolution from severely undersampled game frames, achieving high-fidelity 8$\times$ upscaling from 120p to 960p, with potential for 16$\times$ magnification.
    \item We develop a causal diffusion framework that leverages historical low-resolution frames, in-engine G-buffers, and game-specific adaptation to synthesize high-frequency details at frame rates around 13 FPS.
    \item We introduce a dataset of 336k frames collected from two recent AAA games, all with corresponding G-buffer information in high-resolution.
\end{itemize}

\section{Related Work}
\label{sec:related}

\begin{figure*}[t]
  \centering
   \includegraphics[width=\linewidth]{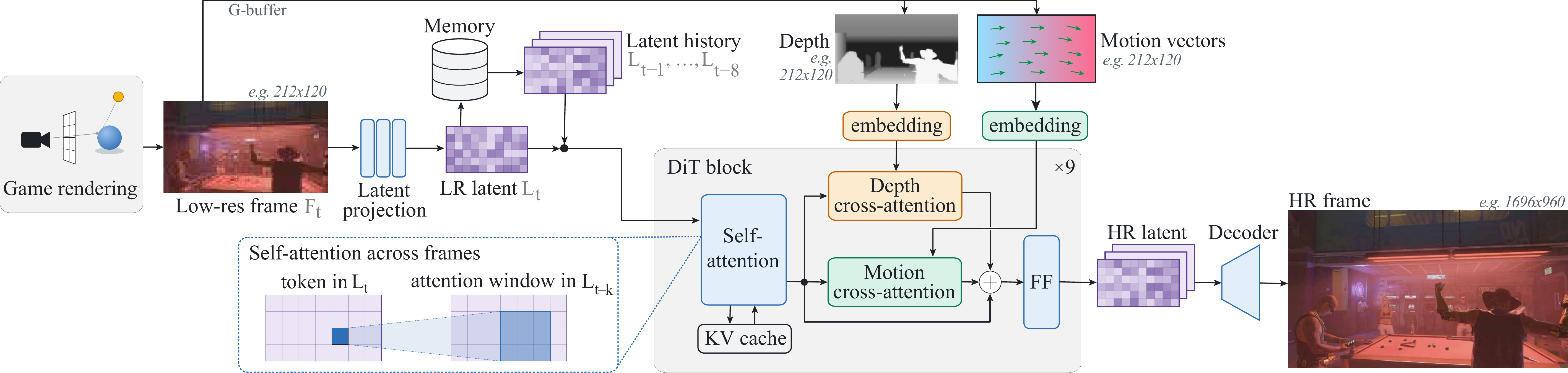}
   \caption{\textbf{Overview of our pipeline}. Each low-resolution frame is projected into latent space and processed by a causal diffusion transformer using cached history and G-buffer conditioning. A lightweight decoder produces the high-resolution frame without future observations or feedback from generated outputs.}
   \label{fig:pipe}
\end{figure*}

\subsection{Video Super-Resolution and Causal Diffusion}
\label{sec:related_vsr_causal}
Video super-resolution (VSR) aggregates information across frames to recover missing details. Early deterministic methods, including recurrent networks~\cite{sajjadi2018frvsr} and bidirectional alignment frameworks~\cite{wang2019edvr, chan2021basicvsr, chan2022basicvsrpp, liang2022rvrt, chan2022realbasicvsr}, perform well at moderate scaling factors but produce overly smooth textures under severe undersampling, as pixel-wise losses average out uncertain high-frequency details~\cite{blau2018perception}. Recent generative VSR methods instead leverage GAN and diffusion priors to synthesize realistic details~\cite{xu2025videogigagan, zhou2024upscaleavideo, li2025diffvsr, wang2025seedvr}. However, standard video diffusion models~\cite{blattmann2023align, blattmann2023stable} use bidirectional temporal attention across frame chunks, requiring future frames unavailable in interactive games where gameplay depends on live player input.

Recent works enable online generation by adapting diffusion models into causal, autoregressive pipelines using rolling KV caches, causal attention masks, and sequential denoising~\cite{yin2025causvid, henschel2024streamingt2v, chen2024diffusion, kim2024fifo}. In VSR, Stream-DiffVSR~\cite{shiu2026streamdiffvsr} and FlashVSR~\cite{zhuang2026flashvsr} support past-frame guidance and low-step streaming inference. Even so, autoregressive video models can suffer from exposure bias and drift over long sequences~\cite{huang2025selfforcing}, and generic video restorers lack explicit scene geometry and motion cues when inputs are heavily undersampled (8$\times$ to 16$\times$). Instead, \acronym maintains causal history over projected low-resolution latents rather than feeding back synthesized high-resolution frames, and anchors generation directly to engine-provided G-buffers (depth and motion vectors), keeping details consistent with the underlying scene without needing future frames.

\subsection{Super-Resolution for Game Rendering}
\label{sec:related_game_sr}
Modern rendering pipelines rely on temporal super-resolution to reduce shading costs while maintaining high output resolution. Commercial upscalers such as NVIDIA DLSS~\cite{nvidia_dlss}, AMD FSR~\cite{amd_fsr}, and Intel XeSS~\cite{intel_xess}, along with academic methods like Neural Supersampling~\cite{xiao2020neural} and ExtraNet~\cite{guo2021extranet}, combine temporal accumulation with engine-generated depth and motion vectors to reproject past frames. Later academic works incorporate richer renderer buffers, such as surface normals and material attributes, for joint denoising and supersampling~\cite{thomas2022temporally}, game-specific reconstruction~\cite{mercier2023gaming}, and radiance demodulation~\cite{li2024nsrd}. Notably, RDG~\cite{zheng2025rdg} processes G-buffers separately from aliased color inputs to better guide temporal feature refinement at up to 4$\times$ magnification.

While these causal, geometry-aware methods perform well up to 4$\times$ scaling, they rely on temporal sub-pixel accumulation and reprojection. At extreme scaling factors (8$\times$--16$\times$, e.g., 120p gameplay), each input pixel spans 64 to 256 output pixels, leaving sub-pixel samples too sparse and disoccluded regions unrecoverable through reprojection alone. Rather than claiming G-buffer conditioning or causal processing as individually novel, \acronym combines both within generative diffusion, using depth and motion vectors to guide high-frequency synthesis where the color input provides insufficient evidence.

\subsection{Generative Priors for Visual Restoration}
\label{sec:related_generative_sr}
When inputs are heavily degraded or upscaled by large factors, deterministic regression produces blurry outputs, making generative priors necessary to synthesize plausible high-frequency detail. In image restoration, GAN-based methods such as BSRGAN~\cite{zhang2021bsrgan} and Real-ESRGAN~\cite{wang2021realesrgan} handle complex degradations, while diffusion-based approaches, including ResShift~\cite{yue2023resshift}, DiffBIR~\cite{lin2024diffbir}, PASD~\cite{yang2024pasd}, SeeSR~\cite{wu2024seesr}, and SUPIR~\cite{yu2024supir}, use learned generative priors to balance realistic texture synthesis with fidelity to the input. For video, VideoGigaGAN~\cite{xu2025videogigagan} shows that high-factor magnification is possible, confirming that extreme upscaling itself is not new. However, standard image and offline video restorers operate without real-time causality or explicit scene constraints. Interactive games require online frame generation where synthesized details stay faithful to scene geometry and motion, which we explore at the highly restrictive 120p input resolution.

\subsection{Efficient Diffusion and Latent Representations}
\label{sec:related_efficient_diffusion}
Latent diffusion models~\cite{rombach2022ldm, chen2025dove} reduce denoising costs by operating in a compressed space, yet VAE encoding, decoding, and iterative denoising remain too slow for interactive pipelines. To speed up inference, distribution matching distillation~\cite{yin2024dmd} enables few- or one-step generation, an approach adopted by recent video restoration models such as FlashVSR~\cite{zhuang2026flashvsr}, alongside prediction-and-refinement methods like PS-SR~\cite{wu2026pssr}. In addition, standard VAE encoders in diffusion upscalers expect high-resolution inputs, requiring low-resolution frames to be upsampled before encoding and incurring unnecessary compute. Because our ultra-low-resolution inputs already match the spatial dimensions of the backbone's compressed latent grid at 8$\times$ scaling, we replace the expensive upsample-and-encode path with a lightweight convolutional projector that maps 120p inputs directly into the latent space, paired with a modified CogVideoX-compatible tiny autoencoder \cite{BoerBohan2025TAEHV} for efficient single-frame decoding. Rather than claiming a net compute advantage over native rendering, our goal is to make high-quality, causal diffusion upscaling from extreme input resolutions computationally feasible.

\section{Method}
\label{sec:method}

\subsection{Overview}

We address causal, frame-by-frame super-resolution of ultra-low-resolution gameplay by adapting the pretrained DOVE \cite{chen2025dove} backbone through three complementary components: a framewise latent pipeline with cached causal attention; transformer and decoder compression combined with inference optimizations; and model distillation followed by pixel-space refinement to recover reconstruction quality. We adopt DOVE for its strong reconstruction performance and straightforward one-step restoration architecture. Unlike noise-conditioned approaches, DOVE directly restores degraded input latents, allowing us to replace its expensive VAE encoder with a lightweight projector while retaining the pretrained transformer. Our primary configuration performs $8 \times$ upscaling from $212 \times 120$ to $1696 \times 960$. The pipeline generates a single HR frame per generation pass, using only the current and past low-resolution inputs. This avoids waiting for additional frames to form a chunk, enabling the low-latency processing needed for interactive gameplay. 


Let $x_t \in \mathbb{R} ^{H \times W \times 3}$ denote the low-resolution (LR) frame at time $t$, with optional engine-provided depth $d_t$ and screen-space motion $v_t$. As shown in \cref{fig:pipe}, for each incoming LR frame $x_t$, we project it into the latent space, use a causal diffusion transformer to predict the corresponding high-resolution (HR) latent conditioned on cache history, and decode a single HR output frame: 
\begin{equation*}
\begin{aligned}
    z_t &= P_{\psi}(x_t), \\
    (\hat{z}_t, C_{t})
    &= F_{\theta}([z_{t-8}, \ldots, z_t],
                   d_t, v_t; C_{t-8:t-1}), \\
    \hat{y}_t &= D_{\phi}(\hat{z}_t).
\end{aligned}
\end{equation*}
where $P_{\psi}, F_{\theta}, D_{\phi}$ denote the LR projector, causal diffusion transformer, and single-frame decoder, respectively. We store the projected latents of the eight most recent LR frames and concatenate them with the current frame's LR latent to form the transformer input. Separately, $C_{a:b}$ denotes the transformer's KV cache for frames $a$ through $b$, inclusive. These two caches avoid repeatedly projecting past LR frames and recomputing their attention keys and values. Only the newly arriving LR frame passes through the projector, and only the current predicted HR latent passes through the decoder. Decoded HR output frames are not fed back into the model, avoiding error accumulation and additional encoding costs.

\subsection{Framewise Causal Super-Resolution}

\paragraph{Direct Latent Projection.}
The original conditioning path of DOVE \cite{chen2025dove} bilinearly upsamples the LR input to the output resolution and encodes it with the pretrained CogVideoX \cite{yang2025cogvideox} VAE encoder (127 million parameters) to obtain the LR input latent. We replace this expensive path with a learned latent projection that maps the LR frame directly to the latent grid, trained to reproduce the encoder's output. The projector is a dilated residual network with approximately 1.2 million parameters, roughly $100\times$ fewer than the original encoder. For $8 \times$ super-resolution, the LR frame and target latent grid have the same spatial dimensions, and therefore we avoid constructing and encoding an intermediate HR image. 

\paragraph{Causal Temporal Context.}
We represent each frame with its own latent and use a temporal patch size of one, preserving a separate spatial token grid for each frame. We replace the backbone's bidirectional temporal attention with a frame-causal mask where current frame tokens can only attend to one another and to the preceding $K=8$ frames. Each transformer block maintains a KV cache for these past frames, evicting the oldest frame's entries as each new frame arrives. We fine-tune the backbone using the same bounded temporal attention window as streaming inference, allowing training to capture the full history depth used at inference. 
To further reduce inference cost, we restrict attention for past frames to local spatial neighborhoods while retaining full spatial attention within the current frame. Specifically, a query at spatial position $p$ in frame $t$ attends to historical tokens within a neighborhood of radius $r=4$ around $p$: 
\begin{align*}
    \mathcal{H}(t, p)=\{(\tau, q):\max(0, t-K) \leq \tau < t, q \in \mathcal{N}_r(p)\},
\end{align*}
where $\mathcal{N}_r(p)$ denotes the local spatial neighborhood on the token grid. The temporal window bounds KV cache size independently of sequence length, and the spatial restriction reduces the number of historical tokens accessed per query. We apply this spatial restriction only at inference; training uses the temporal mask with unrestricted spatial attention. 

\paragraph{Engine-Buffer Conditioning.}
We incorporate depth and two-channel motion through separate attention branches inside each transformer block. Lightweight convolutional encoders (70,000 parameters each) map the normalized buffers to spatial tokens aligned with the video-token grid. Following the standard joint self-attention operation, video features query these representations through local depth and motion cross-attention branches, whose outputs enter as scaled residual updates. 
The buffer encoders operate causally, so processing a frame does not require future engine buffers. 


\paragraph{Single-Frame Reconstruction.}
To reduce decoding overhead, we replace DOVE's original CogVideoX VAE decoder with the CogVideoX-specific variant of the Tiny AutoEncoder (TAE) \cite{BoerBohan2025TAEHV}. Starting from its pretrained 9.8M-parameter decoder, we retain only the final slice of its temporal-growth output and remove historical memory inputs, reducing the decoder to 7.5M parameters and enabling single-frame reconstruction. We then fine-tune the decoder on the transformer's predicted HR latents to adapt it to the output distribution of our model. Since temporal context is already incorporated by the transformer, the decoder reconstructs each frame independently without requiring multi-frame decoding.

\subsection{Model Compression and Efficient Inference}\label{sec:compression}

Our teacher is based on DOVE's 42-block CogVideoX-5B transformer. To reduce inference cost, we construct a compact student with 12 transformer blocks, 24 attention heads, and a hidden dimension of 1536, comprising approximately 378 million parameters including the buffer-conditioning branches. We initialize the student by selecting and transferring teacher weights based on their magnitudes, followed by full-weight knowledge distillation. We additionally fold constant adaptive-normalization modulation into the network and represent fixed text conditioning using learned constant keys and values.

Following decoder adaptation and joint pixel-space refinement (\cref{sec:obj}), we further compress the student through structured block pruning. We rank transformer blocks by the validation degradation caused by removing each block without retraining, prune three low-impact blocks, and fine-tune the remaining network to recover reconstruction quality. The resulting nine-block deployment model contains approximately 291 million parameters while preserving causal KV caching, engine-buffer conditioning, latent projection, and single-frame decoding.

We additionally implement a fused Triton streaming-attention kernel. For each new frame, the kernel combines attention over cached local neighborhoods and constant text tokens with the current-frame attention result, reducing intermediate memory traffic and kernel-launch overhead. Together, with the compilation of the transformer and decoder, these optimizations reduce per-frame execution cost while retaining the streaming attention computation.

\subsection{Training Objectives}\label{sec:obj}

\paragraph{Projector Pretraining and Causal Model Training.} 
We pretrain the LR projector to produce cached posterior-mean latents from the original input-conditioning path using mean-squared error and an L1 loss. Reference latents are obtained by encoding bilinearly upsampled LR frames with the original VAE encoder. We train the backbone using rank-64 LoRA with a scaling parameter of 64, together with the newly introduced conditioning modules. Training progressively incorporates engine buffers, perceptual supervision, per-frame latents, the learned projector, and causal attention.



\paragraph{Latent-Space Distillation.}
We use the trained full-size causal model as a teacher for latent-space knowledge distillation. The objective combines mean-squared error between student and teacher predictions of the clean HR latent, an L1 loss matching their consecutive-frame latent differences, and normalized mean-squared error between selected student and teacher block features. This stage operates entirely in the latent space, avoiding decoder execution and perceptual loss evaluation. 




\paragraph{Decoder Adaptation and Joint Refinement.} We first fine-tune the lightweight decoder on cached student predictions paired with HR targets, while keeping the student fixed. This adapts the decoder to the latent distribution encountered at inference. We then jointly optimize the student and decoder against paired HR sequences without teacher supervision. Both stages use a pixel-space objective combining L1 reconstruction loss, LPIPS, and DISTS \cite{ding2020iqa} with weights of 1, 0.5, and 0.5, respectively, together with a temporal L1 loss matching consecutive-frame differences between predicted and reference sequences. Following block pruning (\cref{sec:compression}), we fine-tune the remaining student and decoder using the same objective to recover reconstruction quality. Joint refinement and pruning recovery preserve the eight-frame causal attention window used during streaming inference.

\section{Experiments}

\newcommand{\redcheck}{{\textbf{\color{red}\XSolidBrush}}}
\definecolor{mygreen}{RGB}{35,226,0}
\newcommand{\greencross}{{\textbf{\color{mygreen}\Checkmark}}}

\begin{table*}[t]
\centering
\caption{\textbf{Quantitative comparison with state-of-the-art VSR methods.} We compare our method against simple bicubic interpolation, our baseline (DOVE), and recent streaming and efficient VSR methods across speed and perceptual/fidelity metrics. Note that \emph{latency} (time required to produce an output from the available input frame(s); chunk-by-chunk generation may encounter higher latency compared with frame-by-frame generation) vs. \emph{ms/frame} (time to render each frame on average) measure two different aspects of the system performance.}
\label{tab:main_baselines}
\setlength{\tabcolsep}{4pt}
\resizebox{\textwidth}{!}{%
\begin{tabular}{l*{14}{c}}
\toprule
\multirow{2}{*}{\textbf{Method}}
& \multirow{1}{*}{\textbf{Causal?}}
& \multirow{1}{*}{\textbf{Per-Frame?}}
& \multicolumn{2}{c}{\textbf{Inference Speed}}
& \multicolumn{5}{c}{\textbf{Cyberpunk}}
& \multicolumn{5}{c}{\textbf{Wukong}} \\
\cmidrule(lr){4-5}
\cmidrule(lr){6-10}
\cmidrule(lr){11-15}
& (w/o relying on future)
& (w/o chunks)
& latency (ms) $\downarrow$
& ms/frame $\downarrow$
& PSNR $\uparrow$
& SSIM $\uparrow$
& LPIPS $\downarrow$
& MILO $\downarrow$
& CGVQM $\uparrow$
& PSNR $\uparrow$
& SSIM $\uparrow$
& LPIPS $\downarrow$
& MILO $\downarrow$
& CGVQM $\uparrow$ \\
\midrule

Bicubic
& \greencross
& \greencross
& \textbf{0.5} & \textbf{0.5}
& 29.37 & 0.8549 & 0.4219 & 0.01070 & 63.20
& 28.95 & 0.8050 & 0.5088 & 0.0115 & \textbf{53.64} \\

DOVE (Original) \cite{chen2025dove}
& \redcheck
& \redcheck
& 10{,}881 & 1{,}209
& 28.13 & 0.8258 & 0.2144 & 0.00490 & 52.08
& 27.22 & 0.7478 & 0.3100 & 0.0058 & 36.01 \\

FlashVSR \cite{zhuang2026flashvsr}
& \redcheck
& \redcheck
& 2{,}642 & 137
& 26.97 & 0.7958 & 0.2609 & 0.00710 & 43.68
& 26.59 & 0.7261 & 0.2919 & 0.0063 & 29.79 \\

Stream-DiffVSR \cite{shiu2026streamdiffvsr}
& \greencross
& \greencross
& 571 & 571
& 29.11 & 0.8471 & 0.3465 & 0.00980 & 56.43
& 28.91 & 0.8013 & 0.4099 & 0.0104 & 49.34 \\

RDG \cite{zheng2025rdg}
& \greencross
& \greencross
& 8.3 & 8.3
& 28.85 & 0.8480 & 0.3850 & 0.00980 & 60.74
& 28.53 & 0.8022 & 0.4819 & 0.0105 & 51.67 \\

InfVSR \cite{zhang2026infvsr}
& \greencross
& \redcheck
& 4{,}667 & 519
& 26.54 & 0.8020 & 0.2720 & 0.00640 & 41.14
& 26.25 & 0.7041 & 0.3748 & 0.0077 & 13.75 \\

DLSS \cite{nvidia_dlss}
& \greencross
& \greencross
& 1.8 & 1.8
& 26.12 & 0.8501 & 0.3023 & 0.01026 & 58.65
& 27.80 & \textbf{0.8125} & 0.3533 & 0.00824 & 51.79 \\

\midrule
\textbf{Ours}
& \greencross
& \greencross
& 75 & 75
& \textbf{31.64}
& \textbf{0.8855}
& \textbf{0.1084}
& \textbf{0.00251}
& \textbf{68.64}
& \textbf{29.76}
& 0.8120
& \textbf{0.1652}
& \textbf{0.0039}
& 50.61 \\

\bottomrule
\end{tabular}%
}
\end{table*}

\subsection{Experimental Setup}
\label{sec:setup}

\paragraph{Dataset.}
Our evaluation requires gameplay footage of state-of-the-art games with ray tracing support and G-buffer recordings.
To the best of our knowledge, no publicly available dataset fulfills both requirements sufficiently.
We therefore record our own dataset from two AAA games, \emph{Cyberpunk 2077} and \emph{Black Myth: Wukong}, using XeSS Inspector. \footnote{Gameplay footage from Cyberpunk 2077
(\copyright{} CD Projekt S.A.) and Black Myth: Wukong
(\copyright{} Game Science) is used solely for
non-commercial academic evaluation. No affiliation
with or endorsement by the respective studios is
implied. Raw game assets are not redistributed.}
\emph{Cyberpunk 2077} data is collected with a simple bot. \emph{Black Myth: Wukong} is collected from human gameplay.
We capture frames at $2560{\times}1440$ with ray tracing enabled at native resolution, without any upscaling or frame generation. Each recording contains 480 frames captured at 30 FPS.
From the G-buffer, we record depth and motion vectors.
For Cyberpunk 2077, we record five distinct scenes (100 videos per scene) and 100 additional videos (20 videos per scene) at $3840{\times}2160$ (for ablation).
Black Myth: Wukong has no distinct scenes, so we treat its 100 videos as a single scene group.
We tone-map the linear HDR frames and resize them with bilinear filtering to $1696 \times 960$ to obtain our HR targets. For the LR inputs, we point sample from the HR frames on a regular grid ($212 \times 120$ for $8 \times$) to mimic rendering at lower resolution.
We apply an 80/10/10 training–validation–test split to the four Cyberpunk scenes used for training and to the Wukong recordings. We additionally reserve the fifth Cyberpunk scene, Santo Domingo, entirely for held-out generalization evaluation (see \cref{sec:abl}).
Unless stated otherwise, we report results for upscaling from $212{\times}120$ to $1696{\times}960$ ($8{\times}$ per axis), averaged over the four-scene Cyberpunk 2077 test set.
Further resolutions, scaling factors, and per-scene results are analyzed in \cref{sec:abl}.

\paragraph{Comparison Methods.}
Beyond simple bicubic interpolation, we compare our method against two groups of baselines.
As an industry-grade real-time upscaler, we evaluate NVIDIA DLSS Super Resolution via NVIDIA Streamline SDK v2.12.0. Since DLSS only supports \(1{\times}, 2{\times}, 3{\times}\) input-to-output ratio in this setting, we obtain \(8{\times}\) enlargement by cascading three $2\times$ DLSS passes: \(212{\times}120 \rightarrow 424{\times}240 \rightarrow 848{\times}480 \rightarrow 1696{\times}960\). At each stage, the previous DLSS output is re-fed as the next-stage color input, while depth and motion vectors are resized to the current input resolution (bilinear interpolation); motion-vector magnitudes are rescaled consistently with the resolution change. We report the output of the third pass as the \(8{\times}\) DLSS baseline. Note that the original DLSS output contains an overlay watermark. Therefore, we evaluated DLSS on only the cropped areas outside the masks.

As state-of-the-art research methods, we evaluate FlashVSR \cite{zhuang2026flashvsr},  Stream-DiffVSR \cite{shiu2026streamdiffvsr}, RDG \cite{zheng2025rdg}, and InfVSR \cite{zhang2026infvsr}.
All baselines receive the same low-resolution inputs as our method and are used with their released weights.
We additionally report DOVE \cite{chen2025dove} as an offline reference to estimate the quality that a neural upscaler can reach without real-time constraints.
DOVE is not applicable to games, as it violates both requirements of the setting: it does not run at interactive frame rates, and it processes whole clips and therefore depends on future frames. For baselines that support only $4 \times$ upscaling, we first bicubically upsample the LR input by $2 \times$ and then apply the model's native $4 \times $ upscaling to obtain an overall $8 \times$ output. We found that this configuration obtains the best performance for these models. 

\paragraph{Evaluation Metrics.}
We measure the quality of the reconstructed frames vs. the rendered ground truth.
We report perceptual quality with LPIPS \cite{zhang2018unreasonable} and MILO \cite{cogalan2025milo}, and pixel-level error with PSNR/SSIM. 
All quality metrics are computed per frame and averaged over all frames of the test set.
CGVQM reports video quality of the full clips in comparison with the reference \cite{jindal2025cgvqm}.
In addition, we measure the inference time of the super-resolution step in milliseconds per frame and convert it to the corresponding theoretical frame rate.
The timings isolate the super-resolution step and do not include ray tracing of the low-resolution input.

\paragraph{Implementation Details.}
The compact student is initialized and distilled as described in (\cref{sec:compression}-\cref{sec:obj}), using full-weight optimization for 11,000 steps at a learning rate of $10^{-4}$, followed by 1,200 steps of joint pixel-space refinement at $2 \times 10^{-5}$. The temporal reconstruction loss is weighted by 0.5. After refining the 12-block student, we prune three transformer blocks and fine-tune the resulting nine-block model to recover reconstruction quality. Training preserves an eight-frame temporal context with banded causal attention; inference additionally restricts historical attention to a local spatial window of radius four. Training takes approximately 72 hours in total on a single NVIDIA H200 GPU. All evaluations and timings are performed on an NVIDIA RTX Pro 6000 Blackwell GPU.
\subsection{Results}

\begin{figure*}[t]
  \centering
   \includegraphics[width=\linewidth]{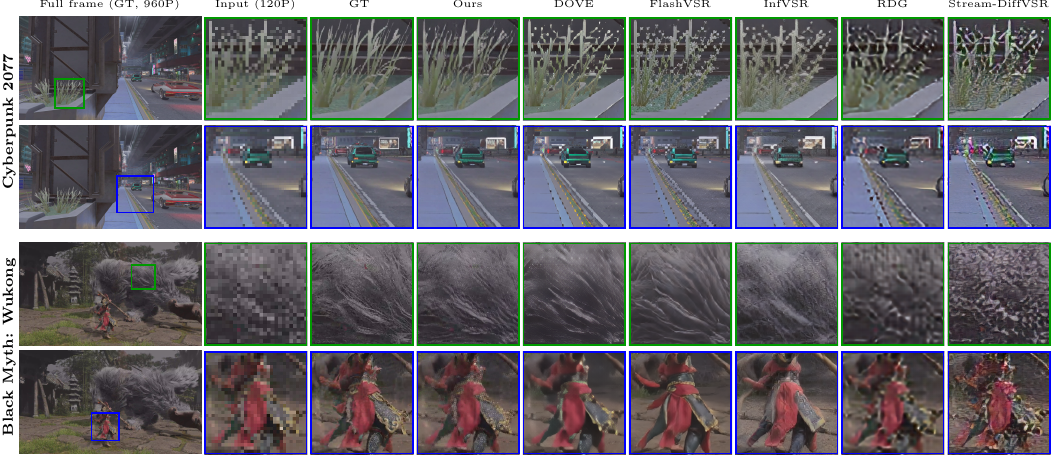}
   \caption{\textbf{Qualitative comparison across the baselines and our approach.} The colored boxes indicate regions for zoomed-in visual inspection. Note either over-smoothing or noisy details among baselines compared with our approach.}
   \label{fig:results}
\end{figure*}

\paragraph{Quantitative Results.}
\Cref{tab:main_baselines} compares \acronym against state-of-the-art video super-resolution methods under extreme 8$\times$ upscaling from 120p inputs. At this resolution, much of the high-frequency information is absent from individual frames, making faithful reconstruction particularly challenging. \acronym achieves the best PSNR, LPIPS, and MILO scores across both games, as well as the highest SSIM on Cyberpunk 2077.
On Cyberpunk 2077, our method achieves 31.64 dB PSNR, exceeding the next-best result by 2.27 dB, while reducing LPIPS by 49\% compared with DOVE. Similar improvements are observed on Wukong, where \acronym reduces LPIPS by 43\% compared with the next-best method.
\acronym also achieves the highest CGVQM on Cyberpunk 2077, although its score on Wukong is slightly below Bicubic, DLSS, and RDG.
These results highlight the advantage of our generative approach in recovering scene details that are difficult to reconstruct through conventional temporal accumulation, while maintaining fidelity to the reference rendering.


\paragraph{Qualitative Results.}
\Cref{fig:results} presents visual comparisons on challenging scenes from both games.
Existing methods struggle with severe undersampling, often producing blurred textures, distorted fine structures, and residual aliasing artifacts.
In contrast, \acronym synthesizes missing high-frequency details while better preserving scene geometry and appearance.
Additional visual and temporal comparisons are provided in the supplementary material.

\begin{table}[t]
\centering
\caption{\textbf{Multi-resolution scaling benchmark.}
Quantitative evaluation of visual and temporal quality and inference
latency across input resolutions and upscaling factors.}
\label{tab:eval_resolution}
\setlength{\tabcolsep}{4pt}
\resizebox{\columnwidth}{!}{%
\begin{tabular}{*{9}{c}}
\toprule
\textbf{Input}
& \textbf{Output}
& \textbf{Scale}
& \textbf{Inference Speed}
& \multicolumn{5}{c}{\textbf{Visual \& Temporal Quality}} \\
\cmidrule(lr){4-4}
\cmidrule(lr){5-9}
\textbf{Res.}
& \textbf{Res.}
& \textbf{Factor}
& ms/frame $\downarrow$
& PSNR $\uparrow$
& SSIM $\uparrow$
& LPIPS $\downarrow$
& MILO $\downarrow$
& CGVQM $\uparrow$ \\
\midrule
60p
& 480p
& $8\times$
& 22
& 30.71 & 0.8639 & 0.1193 & 0.00334 & 59.09 \\

60p
& 960p
& $16\times$
& 73
& 28.09 & 0.8285 & 0.1876 & 0.00584 & 50.18 \\
\midrule
120p
& 480p
& $4\times$
& 23
& 34.23 & 0.9244 & 0.0663 & 0.00090 & 75.79 \\

120p
& 960p
& $8\times$
& 75
& 31.64 & 0.8855 & 0.1084 & 0.00251 & 68.64 \\
\midrule
135p
& 2160p
& $16\times$
& 980
& 30.42 & 0.8604 & 0.2027 & 0.00550 & 62.68 \\
\bottomrule
\end{tabular}%
}
\end{table}


\begin{table}[t]
\centering
\caption{\textbf{Conditioning ablation.}
We evaluate the effect of depth and motion conditioning on inference
speed and visual and temporal quality.
Best values are shown in bold.}
\label{tab:ablation_cond_mem}
\setlength{\tabcolsep}{4pt}
\resizebox{\columnwidth}{!}{%
\begin{tabular}{l*{6}{c}}
\toprule
\multirow{2}{*}{\textbf{Conditioning}}
& \textbf{Inference Speed}
& \multicolumn{5}{c}{\textbf{Visual \& Temporal Quality}} \\
\cmidrule(lr){2-2}
\cmidrule(lr){3-7}
& ms/frame $\downarrow$
& PSNR $\uparrow$
& SSIM $\uparrow$
& LPIPS $\downarrow$
& MILO $\downarrow$
& CGVQM $\uparrow$ \\
\midrule
None (RGB Baseline)
& \textbf{65}
& 31.54 & 0.8847 & 0.1105 & 0.00255 & 68.46 \\

Depth Only
& 69
& 31.56 & 0.8847 & 0.1101 & 0.00256 & 68.47 \\

Motion Only
& 68
& 31.62 & \textbf{0.8855} & 0.1086
& \textbf{0.00205} & \textbf{68.64} \\

Depth + Motion (Ours)
& 75
& \textbf{31.64} & \textbf{0.8855} & \textbf{0.1084}
& 0.00251 & \textbf{68.64} \\
\bottomrule
\end{tabular}%
}
\end{table}


\paragraph{Discussion.}
Beyond visual fidelity, extreme super-resolution for interactive games must operate causally, since future frames depend on player actions and are unavailable at inference time.
Unlike offline video diffusion approaches, \acronym processes each incoming frame using only current and historical observations, guided by engine-provided depth and motion vectors.
This design enables frame-by-frame generation without waiting for future frames or feeding synthesized HR outputs back into the temporal history.
Meanwhile, our efficient latent representation and model compression reduce inference latency to 75 ms per frame (13.3 FPS), compared with 10.9 seconds of initial latency and 1.21 seconds per frame for the original DOVE.
\acronym also achieves lower inference latency than the evaluated diffusion-based VSR baselines.
Although it remains slower than conventional non-generative game upscalers such as RDG (see \cref{sec:limitations} for detailed discussion), \acronym operates in a more challenging reconstruction regime, where missing visual information must be synthesized rather than recovered solely from existing samples.
Together, these results demonstrate the feasibility of causal diffusion for extreme game super-resolution, bridging the gap between high-fidelity generative reconstruction and the latency requirements of interactive rendering.

\subsection{Ablations}\label{sec:abl}

\begin{figure}[t]
  \centering
   \includegraphics[width=\linewidth]{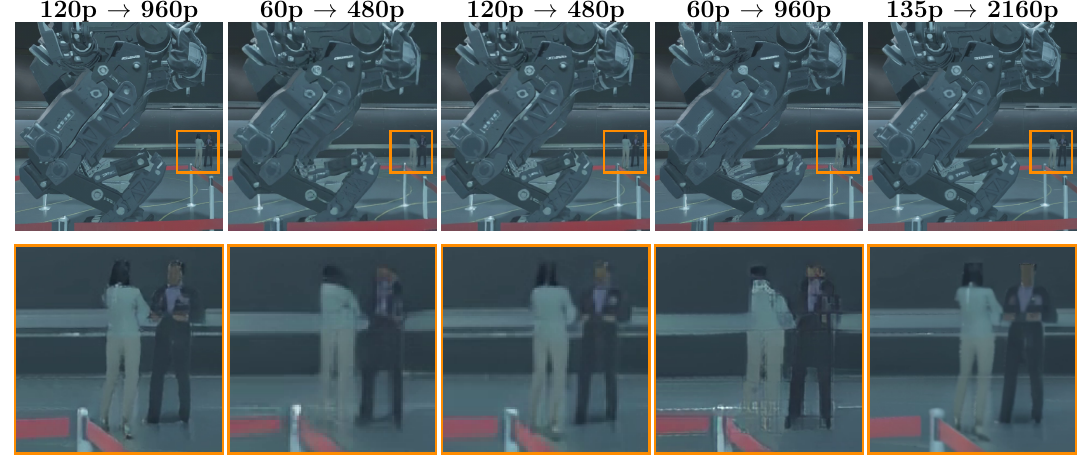}
   \caption{Qualitative comparison across input resolutions and
  upscaling factors, highlighting differences in detail reconstruction
  as input resolution and scaling factor vary.}
   \label{fig:resolutions}
\end{figure}

\paragraph{Scaling Factor and Input Resolution.}
We evaluate the limits of extreme super-resolution by varying input resolution and scaling factor shown in \cref{fig:resolutions}. In \cref{tab:eval_resolution}, increasing the scaling factor from 4$\times$ to 8$\times$ at 120p reduces PSNR from 34.23 to 31.64 dB.
Even at 16$\times$ upscaling from 60p to 960p, \acronym achieves 28.09 dB PSNR and 0.1876 LPIPS, demonstrating its ability to reconstruct visual details from severely undersampled inputs.
Inference latency remains similar for the same output resolution, suggesting that computation is primarily determined by output size.

\paragraph{Engine-Buffer Conditioning.}
\cref{tab:ablation_cond_mem} evaluates depth and motion-vector conditioning.
Compared with RGB-only inputs, incorporating both buffers improves PSNR and LPIPS slightly.
Motion vectors contribute most of the improvement, while depth provides smaller additional gains.
These results suggest that engine-provided information complements generative reconstruction, with a modest latency increase from 65 to 75 ms.

% ...
\nothing{
\paragraph{Acceleration and Model Compression.}
\TODO{\cref{tab:ablation_distill} QS: what's happening with this table/distill??}shows the impact of successive optimizations.
Causal attention with KV caching reduces DOVE's initial latency from 10.88 s to 705 ms, enabling frame-by-frame generation.
Student distillation further reduces latency to 95 ms, and block pruning brings it to 75 ms.
Together, these optimizations achieve a 145$\times$ reduction in initial latency and 16$\times$ faster per-frame processing compared with the original DOVE, making diffusion-based extreme super-resolution more practical for interactive applications.
}

\begin{table}[t]
\centering
\caption{\textbf{Context memory length ablation.}
Effect of the number of past context frames on inference speed
and visual and temporal quality. Best values are shown in bold.}
\label{tab:ablation_context}
\setlength{\tabcolsep}{4pt}
\resizebox{\columnwidth}{!}{%
\begin{tabular}{c*{6}{c}}
\toprule
\multirow{2}{*}{\textbf{Context Frames}}
& \textbf{Inference Speed}
& \multicolumn{5}{c}{\textbf{Visual \& Temporal Quality}} \\
\cmidrule(lr){2-2}
\cmidrule(lr){3-7}
& ms/frame $\downarrow$
& PSNR $\uparrow$
& SSIM $\uparrow$
& LPIPS $\downarrow$
& MILO $\downarrow$
& CGVQM $\uparrow$ \\
\midrule
0
& \textbf{64}
& 30.66 & 0.8725 & 0.1498 & 0.00395 & 66.65 \\

1
& 66
& 31.08 & 0.8764 & 0.1262 & 0.00318 & 67.38 \\

2
& 67
& 31.33 & 0.8807 & 0.1174 & 0.00287 & 68.27 \\

4
& 70
& 31.50 & 0.8832 & 0.1120 & 0.00264 & 68.46 \\

8 (Ours)
& 75
& 31.64 & 0.8855 & 0.1084 & \textbf{0.00251} & 68.64 \\

16
& 84
& \textbf{31.73} & \textbf{0.8869} & \textbf{0.1082}
& 0.00258 & \textbf{68.75} \\

32
& 103
& 30.65 & 0.8689 & 0.1476 & 0.00384 & 63.52 \\
\bottomrule
\end{tabular}%
}
\end{table}
\paragraph{Context Memory Length.}
\cref{tab:ablation_context} shows the tradeoff between temporal context and inference cost.
Increasing memory from 0 to 8 frames improves PSNR from 30.66 to 31.64 dB\nothing{ and reduces LPIPS from 0.1498 to 0.1084}, demonstrating the benefit of past observations for causal reconstruction.
Extending memory to 16 frames offers marginal quality gains at higher latency, while 32 frames degrades quality.
We therefore adopt eight memory frames to balance reconstruction fidelity and inference efficiency.

\paragraph{Held-Out Gameplay Scenes}. 
We evaluate the generalization of \acronym to unseen gameplay scenes in \cref{tab:eval_generalization}.
We train on four visually distinct scenes from Cyberpunk 2077 and evaluate on Santo Domingo, a fifth scene excluded entirely from training.
Despite differences in scenery and visual appearance, \acronym achieves 31.95 dB PSNR and 0.1191 LPIPS on the held-out scene.
These results suggest that \acronym learns transferable game-specific visual priors rather than simply memorizing the training scenes, enabling effective reconstruction in previously unseen environments. This supports the practicality of training \acronym once per game and applying it across diverse in-game environments without scene-specific retraining.

\begin{table}[t]
\centering
\caption{\textbf{Generalization to a held-out Cyberpunk scene.}
Visual and temporal quality on four scenes represented in the
training set and the held-out Santo Domingo scene.}
\label{tab:eval_generalization}
\setlength{\tabcolsep}{4pt}
\resizebox{\columnwidth}{!}{%
\begin{tabular}{l*{5}{c}}
\toprule
\multirow{2}{*}{\textbf{Metric}}
& \multicolumn{4}{c}{\textbf{Training Scenes}}
& \textbf{Held Out Scene} \\
\cmidrule(lr){2-5}
\cmidrule(lr){6-6}
& Streetkid
& Corporate
& Nomad
& Japantown
& Santo Domingo \\
\midrule
PSNR $\uparrow$
& 32.16 & 36.15 & 29.76 & 28.51 & 31.95 \\

SSIM $\uparrow$
& 0.9125 & 0.9560 & 0.8057 & 0.8676 & 0.8969 \\

LPIPS $\downarrow$
& 0.0966 & 0.0722 & 0.1373 & 0.1274 & 0.1191 \\

MILO $\downarrow$
& 0.00209 & 0.00101 & 0.00368 & 0.00326 & 0.00229 \\

CGVQM $\uparrow$
& 74.47 & 85.63 & 53.85 & 60.61 & 70.35 \\
\bottomrule
\end{tabular}%
}
\end{table}

\section{Limitations and Future Work}
\label{sec:limitations}
Despite promising results, \acronym has several limitations.
First, we leveraged game-specific LoRA adaptation for accelerated performance. This limits the scalability to unseen games with different rendering styles, textures, and scene content (see supplementary for more details).
Techniques such as mixture-of-LoRA experts and LoRA merging \cite{shah2024ziplora,liu2026fape} may shed light on improving cross-game generalization by learning shared generative priors across diverse game environments while preserving individual details through lightweight adapters.

Second, although our causal diffusion framework enables extreme super-resolution at around 13 FPS, its inference remains slower than conventional non-generative upscalers. This overhead comes from the computational demands of synthesizing missing high-frequency details rather than reconstructing existing information. Nevertheless, generative upscaling offers opportunities for further efficiency improvements through adaptive computation, compared with fixed-pipeline ray-tracing rendering. For example, future work could selectively allocate generative computation to regions where the low-resolution input provides insufficient detail, while relying on lightweight reconstruction elsewhere. Additionally, jointly optimizing pixel-wise, spatio-temporal ray-tracing samples and generative reconstruction under a shared computational budget could balance visual fidelity and computational cost \cite{hasselgren2020neural}.

\section{Conclusion}
In this work, we present a causal diffusion framework for extreme super-resolution in interactive games, enabling high-quality rendering from ultra-low-resolution inputs as small as 60p. By leveraging generative diffusion priors and G-buffer guidance, our approach synthesizes missing high-frequency details beyond what conventional reconstruction methods can recover. 

\section*{LLM Usage Statement}
We used LLMs to revise and improve the manuscript writing and develop code for part of the algorithm under supervision.
{
    \small
    \bibliographystyle{ieeenat_fullname}
    \bibliography{main}
}


\end{document}